\documentclass{ifacconf}
\usepackage{amsmath,amssymb}
\usepackage{dsfont }
\usepackage{enumerate}
\usepackage{algorithm}
\usepackage{graphicx}      
\usepackage{natbib}        
 
\newtheorem{theorem}{Theorem}[section]

\begin{document}
\begin{frontmatter}

\title{Constrained Reinforcement Learning for Dynamic Optimization under Uncertainty} 

\thanks[footnoteinfo]{EAC was supported by the Engineering and Physical Sciences Research Council (EPSRC) (EP/P016650/1).}

 \author[a]{P. Petsagkourakis}
 \author[b]{I.O. Sandoval}
 \author[c]{E. Bradford}
 \author[d,e]{D. Zhang}
 \author[e]{E.A. del Rio-Chanona}
 \address[a]{Centre for Process Systems Engineering (CPSE), University College London, Torrington Place, London, United Kingdom}
 \address[b]{Instituto de Ciencias Nucleares, Universidad Nacional Aut\'{o}noma de M\'{e}xico, Ciudad de M\'{e}xico, Mexico}
 \address[c]{Department of Engineering Cybernetics, Norwegian University of Science and Technology, Trondheim, Norway}
  \address[d]{Centre for Process Integration, Department of Chemical Engineering and Analytical Science,The University of Manchester, UK}
 \address[e]{Centre for Process Systems Engineering (CPSE), Imperial College London, UK}

\begin{abstract}                
Dynamic real-time optimization (DRTO) is a challenging task due to the fact that optimal operating conditions must be computed in real time. The main bottleneck in the industrial application of DRTO is the presence of uncertainty.
Many stochastic systems present the following obstacles: 1) plant-model mismatch, 2) process disturbances, 3) risks in violation of process constraints. To accommodate these difficulties, we present a constrained
reinforcement learning (RL) based approach.
RL naturally handles the process uncertainty by computing an optimal feedback policy. However, no state
constraints can be introduced intuitively. To address this problem, we present a chance-constrained RL methodology. We
use chance constraints to guarantee the probabilistic satisfaction of process constraints, which is accomplished by
introducing backoffs, such that the optimal policy and backoffs are computed simultaneously. Backoffs are adjusted using the empirical cumulative distribution function to guarantee the satisfaction of a joint
chance constraint. The advantage and performance of this strategy are illustrated
through a stochastic dynamic bioprocess optimization problem, to produce sustainable high-value bioproducts.
\end{abstract}


\end{frontmatter}

\section{Introduction}

The optimization of chemical processes presents a distinctive challenge to the process systems engineering community, given that they suffer from three conditions: 1) there is no precise model known for the process under consideration (plant-model mismatch), leading to inaccurate predictions and convergence to suboptimal solutions, 2) the process is affected by disturbances (i.e. it is stochastic), and 3) process systems can be sensitive, therefore erratic constraint specifications can be inconvenient or even dangerous. 

An efficient dynamic process optimization approach needs to be able to handle both the inherent stochasticity of the system (e.g. process disturbances) and plant-model mismatches, while satisfying safety and physical constraints. To accomplish this, we exploit a method from {\it reinforcement learning} (RL) called {\it policy gradient} with the inclusion of chance constraints from sample approximations. RL has been shown to be a powerful control approach, and is one of the few control techniques able to handle nonlinear stochastic optimal control problems \citep{Bertsekas:2000:DPO:517430}. The inclusion of chance constraints is not new in optimal control, however this work focuses on the development of policies that can handle arbitrary stochastic systems and the constructed policy will require less than a second to be computed. It should further be noticed that  recursive feasibility is out of the scope of this work. Here probabilistic guarantee is provided given the uncertain system. 

Given the rapid development of machine learning technology, building a data-driven model to simulate, optimize, and control complex processes has become possible. In fact, a number of previous studies have adopted supervised learning methods (e.g. artificial neural network, Gaussian processes) to predict process behaviours and conduct open-loop process optimal control \citep{Bradford2018,DelRio-Chanona2017}. However, few studies have been conducted to investigate the applicability and efficiency of reinforcement learning in process engineering, and none of them include the efficient handling of constraints. Therefore, in this work, we propose a policy gradient reinforcement learning algorithm with efficient sample approximation of chance constraints.

Reinforcement learning (in an approximate dynamic programming (ADP) philosophy), has received significant attention for chemical process control. Most of the approaches rely on the (approximate) solution of the Hamilton–Jacobi–Bellman equation (HJBE), and have been shown to be reliable and robust for several problem instances.  For example, in \cite{Lee2005} a model-based strategy and a model-free strategy for control of nonlinear processes were proposed, in \cite{Peroni2005} ADP strategies were used to address fed-batch reactor optimization. In \cite{Tang2018} with the inclusion of distributed optimization techniques, an input-constrained optimal control problem solution technique was presented, among other works (e.g. \cite{Shah2016}). 

In this paper, we present another taking on RL, that of using policy gradient. Policy gradient methods directly estimate the control policy, without the need of a model, or the solution of the HJBE. We highlight their advantages next. In Policy gradient methods, the approximate policy can naturally approach a deterministic policy, whereas action-value methods (that use epsilon-greedy or Boltzmann functions) select a random control action with some heuristic rule. Policy gradient methods work directly with policies that emit probability distributions, which is much faster and does not require an online optimization step. Policy gradient methods are guaranteed to converge at least to a locally optimal policy even in high dimensional continuous state and action spaces, unlike action-value methods where convergence to local optima is not guaranteed. Hence, they also enable the selection of control actions with arbitrary probabilities. In some cases, the best approximate policy may be stochastic \citep{Sutton}. \cite{Petsagkourakis2019a} seems to be the first research where these methods were applied in the context of process optimization and control.

Reinforcement learning and particularly policy gradient methods are considered to be advantageous as discussed above. However, the inclusion of constraints is not straightforward. Various approaches have been proposed in  the literature, where usually penalties are applied for the constraints. Such approaches can be very problematic, easily loosing optimality or feasibility ~\citep{Achiam2017} especially in the case of a fixed penalty. As it is stated in \cite{Wen2018}, existing methods cannot guarantee strict feasibility. The main approaches to incorporate constraints in this way make use of trust-region and fixed penalties~\citep{Achiam2017, Tessler2018}, as well as cross entropy~\citep{Wen2018}. 
Unfortunately, existing methods for constrained reinforcement learning that are based on policy gradient methods cannot guarantee strict feasibility of the output policies even when initialized with feasible initial policies. Also, in \citep{Deisenroth2015, Kamthe2018} a reinforcement learning approach was proposed where the constraints are taken into account but without probabilistic guarantee for the joint constraints. 

To address the above challenges, we propose a method with probabilistic guarantees for the satisfaction of joint chance constraints. We assume a model with uncertainty to be available, with either parametric uncertainty or structural  mismatch. The training of the policy is fully offline and can adapt to different environments as in~\cite{Petsagkourakis2019a}. The proposed method utilizes backoffs for the tightening of the constraints. Several works have been proposed in the area of stochastic MPC including the recently proposed~\cite{koller, Paulson2018, Bradford2019} to account for stochastic uncertainties in NMPC. These methods generally rely on generating closed-loop Monte Carlo (MC) samples offline from the plant to attain the required backoff values.

The structure of this paper is as follows: the problem statement is given in section 2, then in section 3 the details of proposed method for probabilistic satisfaction in reinforcement learning is given. An illustrative case study follows, where the framework is applied in a batch bioreactor. In the last session, the conclusions are discussed.


\section{Problem Statement}
In this work, the dynamic system is assumed to be given by a probability distribution, following a Markov process,
\begin{equation}\label{MC_unc}
    \textbf{x}_{t+1}\sim p(\textbf{x}_{t+1}|\textbf{x}_t, \textbf{u}_t),
\end{equation}
with $\textbf{x}\in \mathbb{R}^{n_x}$, $\textbf{u}\in \mathbb{R}^{n_u}$ and $t$ being the states, control inputs and discrete time, respectively. This behaviour is observed in systems when stochastic disturbances are present and/or other uncertainties affect the physical system, like parametric uncertainties. The case of additive disturbance can be written as:
\begin{equation}\label{additive_unc}
        \textbf{x}_{t+1} =f(\textbf{x}_{t},\textbf{u}_t) + \textbf{w}_t ,
\end{equation}
where $\textbf{w}\in \mathbb{R}^{n_w}$ is a vector of Gaussian distributed additive disturbance. Additionally in the case of parametric uncertainty the model can be described as 
\begin{equation}\label{parametric_unc}
        \textbf{x}_{t+1} =f(\textbf{x}_{t},\textbf{u}_t, \textbf{p}),
\end{equation}
with $\textbf{p}$ being the uncertain parameters. Both (\ref{additive_unc}) and (\ref{parametric_unc}) can be represented by (\ref{MC_unc}).
In this work we seek to maximize an objective function in expectation, using an optimal stochastic policy subject to probabilistic constraints despite the uncertainty of the system. This problem can be written as a Stochastic Optimal Control Problem (SOCP) in (\ref{eq:OCP}). It should be noticed that the notation from reinforcement learning is followed, where the objective is to maximize a total reward instead of minimizing a cost~\citep{Bertsekas:2000:DPO:517430}.

\begin{equation}
\mathcal{P}(\pi^*(\cdot)):=\left
\{\begin{aligned}
        &\max_{\pi(\cdot)} \mathbb{E}[J(\textbf{x}_{t},\textbf{u}_{t})]\\
    &\text{s.t.}\\
    &\textbf{x}_0 = \textbf{x}(0)\\
    &\textbf{x}_{t+1} \sim p(\textbf{x}_{t+1}|\textbf{x}_t, \textbf{u}_t)\\
    &\textbf{u}_t \sim \pi(\textbf{x}_t, D_t)\\
    &\textbf{u}\in\mathbb{U}\\
    &\mathbb{P}(\bigcap_{i=1}^T \{\textbf{x}_i \in \mathbb{X}_i\})\geq 1-\alpha\\
    &  \forall t \in \left\{0,...,T-1\right\}\label{eq:OCP}
\end{aligned}\right.
\end{equation}
with $J$ being the objective function, $\mathbb{U}$ the set of hard constraints for the control inputs and $\mathbb{X}_i$ constraints for states that must be satisfied with a probability $1-\alpha$. Specifically, 
\begin{equation}\label{constraint}
    \mathbb{X}_t = \{\textbf{x}_t\in \mathbb{R}^{n_x}|g_{j, t}(\textbf{x}_t) \leq 0, j =1,\dots,n_g\},
\end{equation} 
and the joint chance constraints are satisfied for the joint event over all $t\in\{1,\dots,T\}$.
Additionally, $\pi(\cdot)$ is the stochastic policy and $D_t$ is a window of past inputs and states that are used by the policy. Unfortunately, this SOCP is intractable in general, and therefore, approximations must be made, some of these approximate methods come from the family of RL algorithms \citep{Bertsekas:2000:DPO:517430, schulman2017proximal, Petsagkourakis2019a}.

In RL a policy $\pi_\theta(\cdot)$ parametrized by the parameters $\theta$, is constructed. This policy, seeks to maximize the expectation of some objective function $J(\cdot)$. We can define this objective function in the finite horizon discrete-time case as
\begin{equation}
    J = \sum_{t=0}^T\gamma^{t}R_t(\textbf{u}_t,\textbf{x}_t),
\end{equation}
where $\gamma\in (0,1]$ is the {\it discount factor} that allows to give more importance to the short-term actions  and $R_t$ a given reward at the time instance $t$ for the values of $\textbf{u}_t$, $\textbf{x}_t$.

The problem that we aim to solve is challenging, since the policy must be
constructed such that it satisfies the probabilistic joint state constraints with some probability, and not only in expectation. The next section discusses the methodology followed by the joint satisfaction of constraints utilizing explicit backoffs, where we can denote a tightened constraint set as:
\begin{equation}\label{constraint1}
    \mathbb{\bar{X}}_t = \{\textbf{x}_t\in \mathbb{R}^{n_x}|g_{j, t}(\textbf{x}_t)+b_{j,t} \leq 0, j =1,\dots,n_g\},
\end{equation} 
where variables $b_{j,t}$ represent the backoffs which tighten the original constraints $\mathbb{X}_t$ defined in (\ref{constraint}).

%
\section{Constrained Policy Optimization for Probabilistic Constraints}
In this section, the general proposed framework for safe reinforcement learning is described. A policy $\pi(\cdot)$ is constructed to optimize in expectation an economic metric of the process ($J$). To accomplish this, a policy optimization is performed, by sampling the physical system (or the generative model) at each time instant. We denote  $\pmb{\tau}$, as the joint random variable of states, controls and rewards for a trajectory with a time horizon $T$:   
\begin{equation}\label{tau}
    \pmb{\tau} = (\textbf{x}_0,\textbf{u}_0,R_0,...,\textbf{x}_{T-1},\textbf{u}_{T-1},R_{T-1},\textbf{x}_T,R_T),
\end{equation}
 then the policy optimization can be defined as:
\begin{equation}\label{policyopt}
    \pi^*_\theta = \arg\max_{\pi_\theta(\cdot)} \mathbb{E}_{\pmb{\tau}\sim p(\pmb{\tau}|\theta)}J(\pmb{\tau}),
\end{equation}
where $p(\pmb{\tau}|\theta)$ represents the probability of the trajectory $\pmb{\tau}$ given the parametrization $\theta$ of the policy. Several approximation are required in  (\ref{policyopt}) to lead to satisfaction of constraints. Recently a methodology was proposed that satisfies the expected value of the constraints~\citep{Tessler2018, Wen2018}, however this is not adequate for most engineering problems, as usually the system is subject to safety constraints, that need to be satisfied with with high probability. To account for constraint violations, in this work, probabilistic constraints are incorporated. The problem is then formulated as:
\begin{equation}\label{policyopt_con}
\begin{split}
    \pi^*_\theta =& \arg\max_{\pi_\theta(\cdot)} \mathbb{E}_{\pmb{\tau}\sim p(\pmb{\tau}|\theta)}J(\pmb{\tau})\\
    &s.t.\\
& \mathbb{P}(\bigcap_{i=1}^T \{\textbf{x}_i \in \mathbb{X}_i\})\geq 1-\alpha\\
& \pmb{\tau} = (\textbf{x}_0,\textbf{u}_0,R_0,...,\textbf{x}_{T-1},\textbf{u}_{T-1},R_{T-1},\textbf{x}_T,R_T)
    \end{split}
\end{equation}
%
We omit the hard constraints for the control inputs in (\ref{policyopt_con}) as they are inherently satisfied by the construction of the policy, e.g. the policy passes through a bounded and differential squashing function~\citep{Deisenroth2015}. 

In order to solve~(\ref{policyopt_con}) three steps must be applied: 1) The policy is parameterized by a multilayer recurrent neural network that computes the mean and variance of the control actions given a state, resulting in a stochastic policy. 2)  The probabilistic constraint in (\ref{policyopt_con}) is substituted by a surrogate set of constraints to guarantee closed-loop probabilistic constraint satisfaction. 3) The `new' constraint is incorporated into the objective function as a penalty to be solved as an unconstrained optimization problem~\citep{Nocedal2006}. The policy optimization is solved by a policy gradient framework~\citep{Sutton:1999:PGM:3009657.3009806}. In the next subsections these components are described.


\subsection{Recurrent Neural Networks}
Recurrent neural networks, (RNNs)~\citep{Rumelhart1986}, are artificial neural networks that have recursive connections between hidden units. This allows them to obtain a `memory' of previous data and model more accurately time-series. Let $\hat{\textbf{x}}_t$ be the vector that contains previous realizations of the states $\textbf{x}$ and the controls $\textbf{u}$, i.e. $\hat{\textbf{x}}_t = \left[\textbf{x}_{t}^T,\dots,\textbf{x}_{t-N}^T,\textbf{u}_{t-2}^T,\dots, \textbf{u}_{t-N-1}^T \right]^T$. Then the stochastic policy can be defined as:
 \begin{equation}\label{rnn}
     \textbf{u}_t \sim \pi_\theta(\textbf{u}_t|\hat{\textbf{x}}_{t},\textbf{u}_{t-1}):=\left
\{\begin{aligned}
         \left[\pmb{\mu}_{t}, \pmb{\Sigma}_{t}\right] &= s_\theta(\hat{\textbf{x}}_{t},\textbf{u}_{t-1})\\
         \textbf{u}_{t} &\sim  \mathcal{N}(\pmb{\mu}_t,\pmb{\Sigma}_t) 
     \end{aligned}\right.,
 \end{equation}
where $s_\theta$ is the multilayer RNN parametrized by $\theta$. Deep structures (which means having more than one hidden layer) are employed to enhance the performance of the learning process ~\citep{Mnih2013,Mnih2015}.  The {\it actual} control inputs are drawn from the mean and variance that has been computed from (\ref{rnn}). Having a stochastic policy could be advantageous when uncertainties are present, as a deterministic policy will always compute the same control inputs with the same states since it learns a deterministic mapping from states to control inputs. On the contrary, a stochastic policy draws a control action from a probability distribution which can account for inherent stochasticity of the uncertain environments. 

%
\subsection{Probabilistic constraints}
In this section, we introduce the surrogate inequality constraints to substitute the probabilistic (chance) constraints. To achieve a probabilistic guarantee, backoff based tightening of individual constraints will be introduced such that satisfaction for the joint chance constraint is attained. 
The probabilistic constraints are intractable, but they can be approximated by the empirical cumulative distribution function (ecdf), using a sample approximation and $S$ Monte Carlo (MC) simulations:
\begin{equation}\label{ecdf}
F = \mathbb{P}(\bigcap_{t=1}^T \{\textbf{x}_t \in \mathbb{X}_t\}) \approx \hat{F}_S = \dfrac{1}{S}\sum_{s=1}^S \mathds{1}\{\bigcap_{t=1}^T \{\textbf{x}^{s}_t \in \mathbb{X}_t\}\ \},
\end{equation}
where $\hat{F}_S$ is the approximate joint constraint satisfaction probability for a trajectory, $\mathds{1}\{\bigcap_{t=1}^T \{\textbf{x}^{s}_t \in \mathbb{X}_t\ \} \} = \begin{cases}1,&  \textbf{x}^{s}_t \in \mathbb{X}_t \, \forall t \in \{ 1,\ldots,T\}\\ 0,& otherwise\end{cases}$. The indicator function is a Bernoulli random variable, which means that $\hat{F}_S$ follows a binomial distribution, with $\hat{F}_S \sim \dfrac{1}{S}\text{Bin}(S, F)$, $F$ being the cumulative distribution function (cdf). The confidence bound for the ecdf can then be computed from the Binomial cdf~\citep{10.2307/2331986}. In fact a simplified expression can be obtained using beta distributions instead leading to the following theorem. 

\begin{theorem}[\cite{10.2307/2331986}]\label{lower}
Assume we are given a value of the ecdf $\hat{F}_S$ (see (\ref{ecdf}))  based on $S$ i.i.d. samples, then the true value of the cdf, $F$, has a lower bound $F_{lb}$ and lies inside the confidence interval $[F_{lb}(\epsilon;S;\hat{F}_S); F_{ub}(\epsilon;S;\hat{F}_S)]$ with a confidence level of $1-\epsilon$. Therefore, the probability of $F$ given a lower bound $F_{lb}(\epsilon;S;\hat{F}_S)$ is as follows:
\begin{equation}
    \begin{split}
        &\mathbb{P}\{F \geq 1-\alpha |{F}_{lb}\geq 1-\alpha\}\geq 1-\epsilon,\\
        &{F}_{lb} = 1-\text{betainv}(1-{\epsilon}, S + 1-S\hat{F}_S, S\hat{F}_S),
    \end{split} 
\end{equation}
with $\text{betainv}(\cdot,\cdot,\cdot)$ being the inverse of the beta cdf with parameters  $\{S + 1-S\hat{F}_S\}$ and $\{S\hat{F}_S\}$, where $1-\alpha$ is the value for the probabilistic constraint satisfaction and $1-\epsilon$ is the confidence level.
\end{theorem}

The objective now is to tighten the individual constraints using backoffs such that the probabilistic lower bound of the ecdf of the joint constraints is equal to $1-\alpha$ (see (\ref{lb_e})), given the confidence level of $1-\epsilon$. 
\begin{equation}\label{lb_e}
    F_{lb} = 1-\alpha.
\end{equation}
In other words, if the lower bound of the ecdf ($F_{lb}$) is equal to $1-\alpha$, then the probability of satisfaction of the joint chance constraints in ($\ref{policyopt_con}$) is larger than $1-\alpha$ with a certainty no smaller than $1-\epsilon$. 
%
Hence, we aim to compute the tightened constraints by backoffs $b_{j,t}$ such that (\ref{lb_e}) is satisfied when the policy optimization has finished.
To find $b_{j,t}$, we first compute an initial set of backoffs ($b_{j,t}^0$), as it has been proposed in~\cite{Paulson2018}, where 
\begin{equation}
    \bar{g}_{j,t}(\textbf{x}_t) + b^0_{j,t} =0~ \forall j,t~\text{gives}~ \mathbb{P}(\textbf{x}_t \in \mathbb{X}_t)\geq 1-\delta, 
\end{equation}
with $\delta$ being a tuning parameter. Additionally, the mean value for $g_{j,t}$ is  the sample average approximation (SAA): $\bar{g}_{j,t} =\frac{1}{S} \sum_{s=1}^Sg_{j,t}^{s}$.
The initial backoff values are computed to probabilistically satisfy each constraint:
\begin{subequations}
\begin{align}
 &\mathbb{P}(\textbf{x}_t \in \mathbb{X}_t)\geq 1-\delta\label{prob_in}\\
    &b^0_{j,t} = \hat{F}_S^{-1}(1-\delta) - (\bar{g}_{j,t}(\textbf{x}_t)),~\forall j,t
\end{align}
\end{subequations}
with $\hat{F}_S^{-1}(1-\delta)$ being the inverse ecdf of (\ref{prob_in}). 
We wish the backoffs $\textbf{b}$ (with elements $b_{j,t}~\forall~j,t~\in \{1,\dots,n_g\}\times\{1,\dots,T\}$) to be large enough to ensure the probabilistic satisfaction of constraints. However, if the backoffs are too large, unnecessary
conservatism will be present, since relaxing any active state constraint can only result in improved performance. Therefore, we wish for all active constraints to be as close to zero as possible. This results in a root-finding problem using Theorem~\ref{lower} with  $F_{lb}-(1-\alpha)= 0$. Given that we do not have the derivatives with respect to the backoffs, we solve this problem by a bisection algorithm. We implement this by iterating over a design parameter $\gamma$, such that $F_{lb}-(1-\alpha) = 0$, and $b_{j,t} = \gamma~b^0_{j,t}$ is used as an update rule. Notice that now $F_{lb}$ is a function of the design parameter $\gamma$, since the lower bound depends on the tightening that is applied.

With the above procedure we are able to compute a surrogate for the probabilistic constraints.


\subsection{Penalty Function and policy gradient method}
In this work we reformulate the constraint using a quadratic penalty function and then solve the problem using a policy gradient method.
A smooth quadratic penalty is added to the objective function of (\ref{policyopt_con})
\begin{equation}
\begin{split}
          \max_{\pi_\theta(\cdot)} \mathbb{E}_{\pmb{\tau}}\left(J(\pmb{\tau}) - \mu \sum_{j=1}^{n_g}\sum_{t=1}^T\max({g}_{j,t}(\textbf{x}_t)), 0)^2\right),
          \end{split} 
\end{equation}
where ${g}_{j,t}$ is given in~(\ref{constraint}) and $\pmb{\tau}$ in (\ref{tau}). It should be noted that the same framework can be implemented in most policy optimization methods \citep{Tessler2018, Achiam2017}. As it is observed in~\cite{Achiam2017}, when penalty methods are applied in policy optimization, depending on the value of parameter $\mu$ the behaviour of the policy may change. If a large value of $\mu$ is used, then the policy tends to be over-conservative resulting in feasible areas that are not optimal; on the other hand, when the value for $\mu$ is too small, the policy tends to ignore the constraints as in the unconstrained optimization case. 
Therefore, the value of $\mu$ must be carefully chosen, with the constraints substituted by the tightened constraints using backoffs. In this way, the policy optimization will compute the `near'-optimal solution, at the same time guaranteeing the probabilistic satisfaction of the constraints for a given value of $\mu$.

\subsection{Proposed Algorithm}\label{alg}

In this work we propose the use of policy gradient, and particularly of the Reinforce algorithm. Reinforce ~\citep{Sutton:1999:PGM:3009657.3009806} approximates the gradient of the policy to maximize the expected reward with respect to the parameters $\theta$ without the need of a dynamic model of the process. It should be mentioned that different algorithms may be applied \citep{schulman2017proximal, Kakade2002}. However the focus here is on the probabilistic guarantee of the constraints. We now define our reward function as:

$\hat{J}(\pmb{\tau}, \textbf{b}) = J(\pmb{\tau}) + \mu \sum_{j=1}^{n_g}\sum_{t=1}^T\max({g}_{j,t}(\textbf{x}_t)) + b_{j,t}, 0)^2$, which is now an explicit function of our probabilistic constraints. 
We now use the Policy Gradient theorem \citep{Sutton:1999:PGM:3009657.3009806} to obtain an explicit gradient of the reward with respect to the parameters that parameterize our policy:
\begin{align}
\nabla_\theta \mathbb{E}_{\pmb{\tau}} (J) \approx \frac{1}{K} \sum_{k=1}^{K}  \left[ \hat{J}(\pmb{\tau}^{k},\textbf{b}) \nabla_\theta \sum_{t=0}^{T-1} \text{log}\left( \pi(\textbf{u}_t^k|\hat{\textbf{x}}_t^k,\theta)\right)\right]
\end{align}
where we denote the sample $k$ as a superscript that denotes the $k^{th}$ sampled trajectory. The variance of this estimation can be reduced with the aid of an action-independent baseline $\bar{\beta}_s$, which does not introduce a bias \citep{Sutton}. A simple but effective baseline is the expectation of reward under the current policy, approximated by the mean over the sampled paths:
\begin{equation}\label{baseline}
    \bar{\beta}_s = \bar{J}(\theta) \approx \frac{1}{S} \sum_{k=1}^K\hat{J}(\pmb{\tau}^{k},\textbf{b}),
\end{equation}
which leads to:
\begin{equation}\label{Rew-Baseline}
	\nabla_\theta \hat J(\theta) \approx \frac{1}{K} \sum_{k=1}^{K}  \left[ (\hat{J}(\pmb{\tau}^k,\textbf{b})-\bar{\beta}_s) \nabla_\theta \sum_{t=0}^{T-1} \text{log}\left( \pi(\textbf{u}_t^k|\hat{\textbf{x}}_t^k,\theta)\right)\right]
\end{equation}
This selection increases the log likelihood of an action by comparing it to the expected reward of the current policy. Equation (\ref{Rew-Baseline}) is the gradient that is used to update the policy in a gradient ascent fashion in policy gradient methods. For fixed values of backoffs, the policy gradient method is outlined in Algorithm~\ref{alg:PG alg1}. 
\begin{algorithm}[H]
\caption{Policy gradient for fixed backoff}\label{alg:PG alg1}
\smallskip
{\bf Input:} Initialize policy parameter $\theta = \theta_0$, with $\theta_0\in\Theta_0$, 
learning rate $\alpha_0$ and its update rule, the number of episodes $K_0$ and the number of epochs $N_0$, $\mu$ parameter, tolerance $tol$, set backoffs $\textbf{b}$\\
{\bf Output:} policy $\pi^{*}(\cdot | \cdot ,\theta)$ and $\Theta$.
\smallskip

{\bf for} m = 1,\dots, N {\bf do} 
\begin{enumerate}
\item Collect $\textbf{u}_t^k , \textbf{x}_t^k$ for $T$ time steps for $K$ trajectories along with $\hat{J}(\textbf{x}_T^k)$, also for $K$ trajectories.
\item Update the policy: $\theta_{m+1} = \theta_m +  \frac{\alpha_m}{K} \sum_{k=1}^{K}  \left[ (\hat{J}(\pmb{\tau}^k, \textbf{b})-\bar{\beta}_s) \nabla_\theta \sum_{t=0}^{T-1} \text{log}\left( \pi(\textbf{u}_t^k|\hat{\textbf{x}}_t^k,\theta)\right)\right]$
\item $history(m+1) := \mathbb{E}(\hat{J})$
\item if $|history(m+1)-history(m)|\leq tol$ then $exit$
\end{enumerate}
\end{algorithm}
Algorithm~\ref{alg:PG alg1} is the base for Algorithm~\ref{alg:PG alg2}, which is discribed next. In \textbf{step (1)}  the policy is trained with $\textbf{b} = 0$, with the initial estimate of the backoffs $b_{j,t}^0$ computed in \textbf{step (2)} with i.i.d. $S$  samples.  Then, the backoff values will repeatedly change (\textbf{3}) until the desired performance is achieved, which corresponds to the satisfaction of (\ref{lb_e}). The value of $\gamma_m$ is the half of $a_m$ and $c_m$ in \textbf{step 3i}. After construction of the new policy, (\ref{lb_e}) is evaluated in \textbf{step 3iv} using $S$ i.i.d. samples and the relevant changes of $a_m$ and $c_m$ subject to bisection method are made in \textbf{ step 3v}. Lastly the algorithm terminates when a desirable tolerance $tol_0$ has been achieved (\textbf{step 3vi}).
It should be noted that every time a new backoff is computed then the policy is re-optimized. This may look inefficient at first glance, however the convergence is achieved fast, as everytime the previous policy is used as an initial guess for the next iteration. In the end of Algorithm \ref{alg:PG alg2}, a probabilistically constrained policy will have been constructed.
\subsection{Policy initialization}
Reinforcement learning (including policy gradient methods) is computationally expensive; this is mainly because most of the the computational cost is shifted offline, but also because initially the agent (or controller in our case) explores the control action space randomly. In the case of process optimization and control, it is possible to use a preliminary controller, along with supervised learning to hot-start the policy, and significantly speed-up convergence. 

The main idea here is to have data from some policy or state-feedback control (e.g. PID controller, (economic) model predictive controller) to compute control actions given observed states. The initial parameterization for the policy (before {\textbf{ step 1}}) is trained in a supervised learning fashion where the states are the inputs and the control actions are the outputs. Subsequently, this parameterized policy is used to initialize the policy in \textbf{step 1} and then trained by the RL algorithm.



\begin{algorithm}[H]
\caption{Backoff-Based Policy Optimization}\label{alg:PG alg2}
\smallskip

{\bf Input:} Initialize policy parameter $\theta = \theta_0$, with $\theta_0\in\Theta_0$, 
learning rate, its update rule $\alpha$, $m:=0$, the number of episodes $K_0, K_1$ and the number of epochs $N_0, N_1$, $\mu$ parameter, set backoffs $\textbf{b} = \textbf{0}$, $\delta<1$, $\alpha< 1$, $tol_0 = 1\times10^{-4}$, $a_0=0$, $c_0$, maximum number of backoff iterations $M$ and $S$ number of samples to compute $\hat{F}_S$\\
{\bf Output:} policy $\pi^{*}(\cdot | \cdot ,\theta)$ and $\Theta$.
\smallskip
\begin{enumerate}
    \item { Perform policy optimization} with $\textbf{b} = \textbf{0}$ using Algorithm~\ref{alg:PG alg1} with $K_0$ episodes and $N_0$ epochs.\\
    Obtain policy $\pi_0(\cdot|\cdot) = \pi^{*}(\cdot|\cdot,\theta)$ 
 \item Estimate initial backoff using $S$ i.i.d. samples:\\ $b^0_{j,t} = \hat{F}_S^{-1}(1-\delta) - \bar{g}_{j,t}(\textbf{x}_t)~\forall j, t$
\item {\bf for} m = 0,\dots, $M$ {\bf do}.
\begin{enumerate}[i]
\item Set $\gamma_m = \dfrac{a_m + c_m}{2}$
\item Set $\textbf{b} = \gamma_m \textbf{b}^0$
    \item { Perform policy optimization} with backoffs $\textbf{b}$ using Algorithm~\ref{alg:PG alg1} with $N_1$, $K_1$ and $tol_0$.\\
    Obtain policy $\pi_\text{m}(\cdot|\cdot) = \pi^*(\cdot|\cdot,\theta)$.
    \item Compute  $e_m = F_{lb} - (1-\alpha)$ using $S$ i.i.d. samples.
    \item if $e_m<0$ then: $a_{m+1} :=\gamma_m$\\ else $c_{m+1} := \gamma_m$
    \item if $|e_m|\leq tol_0$ then {\it exit}
\end{enumerate}
\end{enumerate}
%
\end{algorithm}

%
%
%
%
%
%
\section{Case Study}
The case study in this paper focuses on the photo-production of phycocyanin synthesized by cyanobacterium \textit{Arthrospira platensis}. Phycocyanin is a high-value bioproduct and its biological function is to enhance the photosynthetic
efficiency of cyanobacteria and red algae. It has applications as a natural colorant to replace other toxic synthetic pigments in both food and cosmetic production. Additionally, the pharmaceutical industry considers it as beneficial because of its unique antioxidant, neuroprotective, and anti-inflammatory properties.

\begin{figure}
    \centering
    \includegraphics[scale=0.3]{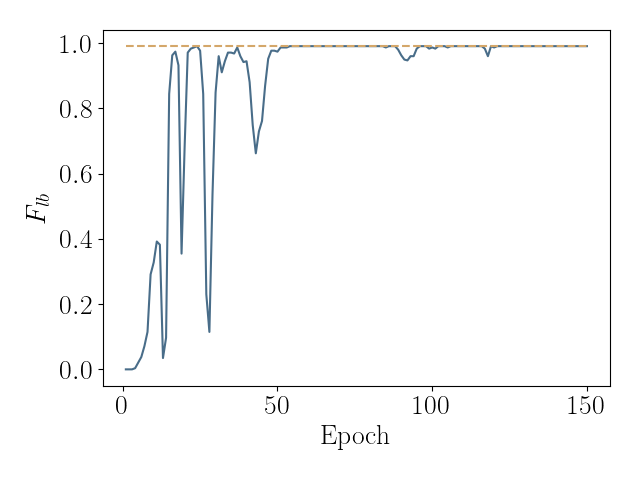}
    \caption{Convergence of lower bound of $F_{lb}$ to $1-\alpha$.}
    \label{fig:lb}
\end{figure}

\begin{figure}
    \centering
    \includegraphics[scale=0.3]{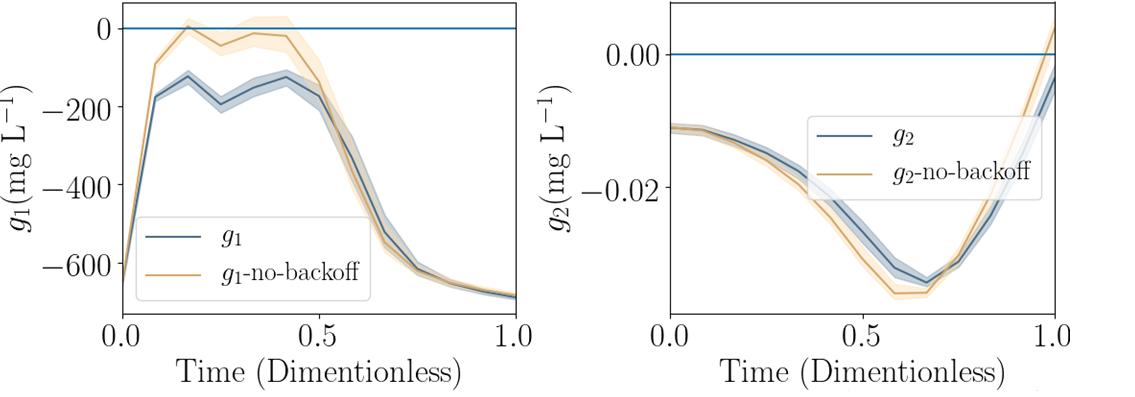}
    \caption{Constraint satisfaction under the absence and presence of backoffs, where the shaded areas are the 98\% and 2\% percentiles.}
    \label{fig:h2}
\end{figure}

\begin{figure}
\centering
 \includegraphics[scale=0.3]{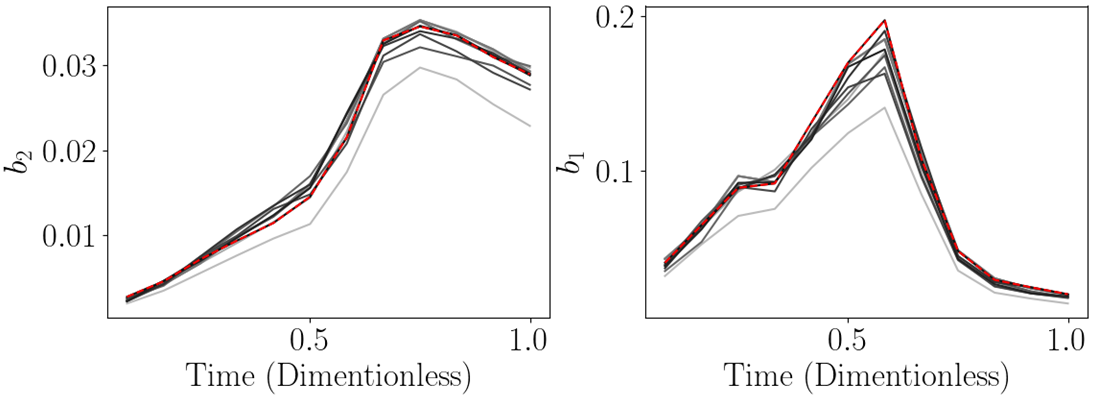}
\caption{The backoff values (dimensionless) are plotted over number of epochs, which are faded out towards earlier iterations.}
 \label{fig:backoff2}
\end{figure}

The  dynamic system consists of the following system of ODEs describing the evolution of the concentration of biomass ($X$), nitrate ($N$), and product ($q$) under parametric uncertainty. The dynamic model is based on Monod kinetics, which describes microorganism growth in nutrient sufficient cultures, where intracellular nutrient concentration is kept constant because of the rapid replenishment. We assume a fixed volume fed-batch. The manipulated variables as in the previous examples are the light intensity ($u_1=I$) and inflow rate ($u_2=F_N$). The mass balance equations are
\begin{align}
     \dfrac{dc_x}{dt} &=  u_m \dfrac{I}{I + k_s + I^2/k_i}c_x \dfrac{c_N}{c_N + K_N} - u_dc_X\label{biomas}\\
     \dfrac{dc_N}{dt} &= -Y_{N/X} u_m \dfrac{I}{I + k_s + I^2/k_i}c_x \dfrac{c_N}{c_N + K_N} + F_N \label{Nitrogen}\\
     \dfrac{dc_q}{dt} &=
      k_m \dfrac{I}{I + k_{sq} + I^2/k_{iq}}c_x \dfrac{c_N}{c_N + K_N} - k_d\dfrac{c_{q}}{C_N + K_{N_q}}       \label{product}
\end{align}
The parameter values are adopted from~\citep{Bradford2019}. Uncertainty is assumed for the initial concentration, where
\begin{align}
    &\begin{bmatrix}
    c_x(0) & c_N(0)
    \end{bmatrix} \sim \mathcal{N}(    \begin{bmatrix}
    1. & 150.
    \end{bmatrix}, diag(1\times 10^{-3}, 22.5)).\\
    &c_q(0) = 0.
\end{align}
Additionally, the 10\% parametric uncertainty of the system:        ${k_s} \sim \mathcal{N}(178.9, 17.89)$, ${k_i} \sim \mathcal{N}(447.1, 44.71)$ and ${k_N} \sim \mathcal{N}( 393.1, 39.31).$
The objective function(reward) in this work is to maximize the product's concentration ($c_q$) at the end of the batch. The objective is additionally penalized by the change of the control actions $\textbf{u}(t) = \left[ I, F_N\right]^{T}$. As a result the reward is given as:
 \begin{equation}\label{reward_bio}
 \begin{split}
     &R_t = -\Delta \textbf{u}_t^T diag(3.125\times10^{-8}, 3.125 \times 10^{-6})  \Delta \textbf{u}_t,\\
     & t\in \{0, T-1\}, R_T = c_q(T),     
 \end{split}
 \end{equation}
 where $\Delta \textbf{u}_t = \textbf{u}_t - \textbf{u}_{t-1}$. The relevant constraints in this work for each time step are $g_{1,t}=c_N - 800\leq 0$ and $g_{2,t}=c_q -0.011c_X\leq 0$. This constraints have been normalized as:
 \begin{equation}
         \tilde{g}_{1,t} =\dfrac{c_N}{800} - 1\leq 0, ~~~
         \tilde{g}_{2,t} = \dfrac{c_q}{0.011c_X} - 1\leq 0
 \end{equation}
and the constraints are meant to be satisfied with probability $99\%$ ($\alpha = 0.01$) and confidence level is $99\%$ ($\epsilon = 0.01$). The constraints are added as a penalty with $\mu = 10$. The control actions are constrained to be in the interval $0\leq F_N \leq 40$ and $120 \leq I\leq 400$, these constraints are considered to be hard. The control policy RNN is designed to contain 4 hidden layers, each of which comprises 20 neurons embedded by a leaky rectified linear unit (ReLU)  activation function. A unified policy network with diagonal variance is utilized such that the control actions share memory and the previous states are used from the RNN (together the current measured states). The  computational cost for each control action online is insignificant since it only requires the evaluation of the corresponding RNN. First the algorithm computes the policy for the bakcoffs to be zero ($\textbf{b} =0$), then the backoffs are updated according to the Algorithm~\ref{alg:PG alg2}. The parameters for the trainings are: $N_0 = 500$, $N_1 = S = 500$, $M = 100$, $K_0 = 150$, $K_1=50$, $tol=tol_0=10^{-4}$ and the two previous states and controls are used from the policy. After the completion of the training the backoffs have been computed and in Fig.~\ref{fig:lb} the convergence of the lower bound $F_{lb}$ is shown. In a rather small number of iterations the backoff values managed to force the $F_{lb}$ to 0.99. The figure shows $F_{lb}$ against all the iterations performed in {\textbf{step 3}} including the training of policy in {\textbf{step 3iii}}.
Now, the actual value of the constraints can be depicted in Fig~\ref{fig:h2}, where the shaded areas are the 98\% and 2\% percentiles. The results are also compared with the case of the absence of backoffs ($\textbf{b}=0$). As it was expected the use of backoffs managed to remain feasible and in the case of the $g_2$ steer the problem to its boundary. 
The normalized backoff values of the $\tilde{g}_{1,t}$ and $\tilde{g}_{2,t}$ for each update are shown in Fig \ref{fig:backoff2}, where the red-dashed represents the converged final value.
Based on the comparison, it is concluded that integrating backoffs into RL can significantly improve RL's optimal control performance when handling complex systems with high stochasticity. From Fig~\ref{fig:h2}, it is seen that the current strategy can well satisfy all the practical constraints; while the RL without backoffs breaches $g_1$ in the middle of the operation and violates $g_2$ at the later stage of the process, hence resulting in an infeasible optimal policy.
It should be noted that the final value for the product $c_q$~(\ref{reward_bio}) is 0.159 and 0.172 when backoffs are applied and when they are not. This difference makes sense since the case of $\textbf{b} = 0$ does not have probabilistic guarantee for the satisfaction of the constraints and the objective can attain higher value in the infeasible area. 

\begin{figure}
    \centering
    \includegraphics[scale =0.3]{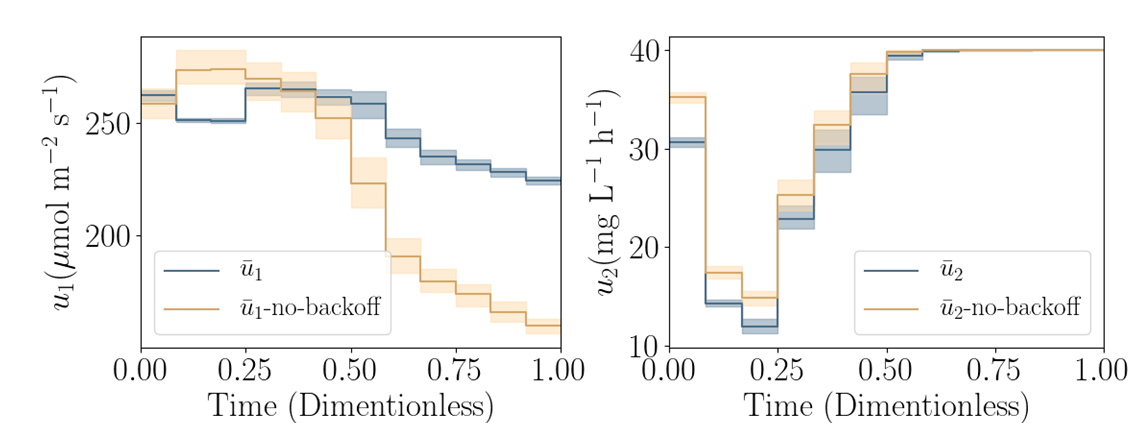}
    \caption{Comparison of the time trajectories of the piecewise constant control actions, where the shaded areas are the 98\% and 2\% percentiles.}
    \label{fig:my_label}
\end{figure}
The algorithm is implemented in Pytorch version 0.4.1.  Adam~\citep{Kingma2014} is employed to compute the network's parameter values using a step size of $10^{-2}$ with the rest of hyperparameters at their default values. 
\section{Conclusions}
 For fermentation and pharmaceutical processes, even a transitory violation of hard constraints may directly damage the product quality and result in an early termination of ongoing batch operation. As a result, choosing a robust online optimization method is of critical importance when uncertainty is present in a process.
The current results further reveal that it is possible to obtain a near optimal and feasible policy given a general uncertain system. In real systems with the absence of a true model, it is impossible to generate highly accurate datasets to train the policy network. As a result, a method that estimates offline backoffs should be thoroughly investigated as it can offer additional advantages to guarantee the probabilistic feasibility of hard constraints when computing an optimal policy. In terms of future work, experimental verification will be conducted to test the efficiency of this strategy and provide suggestions to further improve the current method.

\bibliography{ifacconf}             

\end{document}